\documentclass{article}
\pdfoutput=1

\usepackage{arxiv}
\usepackage[utf8]{inputenc} 
\usepackage[T1]{fontenc}    
\usepackage{url}            
\usepackage{booktabs}       
\usepackage{amsfonts}       
\usepackage{nicefrac}       
\usepackage{microtype}      
\usepackage{lipsum}
\usepackage{graphicx}
\usepackage{xcolor}
\usepackage{enumitem}
\usepackage{amsmath}
\usepackage{booktabs, graphicx, makecell, siunitx, eqparbox}
\usepackage{lineno}
\usepackage{todonotes}
\usepackage{multirow} 

\title{Smooth and Sparse Latent Dynamics in Operator Learning with Jerk Regularization}

\author{
 Xiaoyu Xie \\
    Department of Mechanical Engineering\\
    Northwestern University\\
    Evanston, IL 60208 \\
    \texttt{xiaoyuxie2020@u.northwestern.edu} \\
   \And
 Saviz Mowlavi \thanks{Corresponding author: mowlavi@merl.com} \\
    Mitsubishi Electric Research Laboratories \\
    Cambridge, MA 02139 \\
    \texttt{mowlavi@merl.com} \\
    \And
  Mouhacine Benosman \\
    Mitsubishi Electric Research Laboratories \\
    Cambridge, MA 02139 \\
    \texttt{benosman@merl.com} \\
}

\begin{document}
\maketitle
\begin{abstract}
Spatiotemporal modeling is critical for understanding complex systems across various scientific and engineering disciplines, but governing equations are often not fully known or computationally intractable due to inherent system complexity. Data-driven reduced-order models (ROMs) offer a promising approach for fast and accurate spatiotemporal forecasting by computing solutions in a compressed latent space. However, these models often neglect temporal correlations between consecutive snapshots when constructing the latent space, leading to suboptimal compression, jagged latent trajectories, and limited extrapolation ability over time. To address these issues, this paper introduces a continuous operator learning framework that incorporates jerk regularization into the learning of the compressed latent space. This jerk regularization promotes smoothness and sparsity of latent space dynamics, which not only yields enhanced accuracy and convergence speed but also helps identify intrinsic latent space coordinates. Consisting of an implicit neural representation (INR)-based autoencoder and a neural ODE latent dynamics model, the framework allows for inference at any desired spatial or temporal resolution. The effectiveness of this framework is demonstrated through a two-dimensional unsteady flow problem governed by the Navier–Stokes equations, highlighting its potential to expedite high-fidelity simulations in various scientific and engineering applications.
\end{abstract}


\section{Introduction}

Modeling the dynamics of continuous spatiotemporal systems is a cornerstone of research in fields as diverse as climatology \cite{bi2023accurate} or fluid mechanics \cite{brunton2020machine}, across the physical and biological sciences \cite{alber2019integrating}, as well as in various engineering disciplines \cite{mozaffar2022mechanistic}. The fidelity of such models traditionally relies on the derivation of time-dependent Partial Differential Equations (PDEs) that govern these systems. However, distilling these systems into compact PDE forms is often unfeasible because of complex physics or insufficient knowledge \cite{brunton2016discovering, xie2022data}. Even when governing PDEs are known, obtaining analytical solutions can be impractical, prompting reliance on numerical methods like the finite difference, finite element, or finite volume techniques. While these methods are precise and efficient, they can be computationally intensive, thus impeding high-fidelity simulations, especially in applications requiring iterative simulations such as design optimization, control systems, inverse problems, and uncertainty quantification. This computational intractability, coupled with inherent modeling uncertainties, underscores the significance of developing efficient forecasting models that can aptly capture system behavior directly from existing data, yet return predictions for unseen scenarios at reduced computational overheads.

Data-driven spatiotemporal forecasting models borrowing tools from supervised deep learning have shown promise in predicting future system states. One class of methods employs standard deep learning architectures such as Convolutional Neural Networks (CNNs) \cite{ren2022phycrnet,rao2023encoding} or Graph Neural Networks (GNNs) \cite{pfaff2020learning} for surrogate modeling. These methods, however, are limited because they only map between fixed, finite-dimensional spaces, necessitating that the state be discretized on a preset spatial grid. In contrast, another class of methods termed operator learning techniques, such as the Fourier Neural Operator \cite{li2020fourier}, DeepONet \cite{lu2021learning,pathak2022fourcastnet}, and others \cite{seidman2022nomad,tripura2023wavelet,kissas2022learning}, learns mappings between continuous function spaces, which enables discretization-invariant state predictions that are easily scalable to high-resolution. Despite their differences, both classes of models often share a common limitation: their dependence on autoregressive time-stepping \cite{brandstetter2022message}, which sequentially predicts future system states by using the previous state as an input. This method is computationally intensive, as each state is represented by a discretized high-dimensional vector, making long-term trajectory computation resource-heavy.

The dynamics of spatiotemporal systems often evolve on a manifold of much lower dimension than that of the high-dimensional discretized state \cite{cohen2015approximation,seidman2022nomad}. Taking advantage of this property, a third class of data-driven forecasting methods, Reduced-Order Models (ROMs), seeks to evolve a low-dimensional latent vector of coordinates within this manifold, instead of advancing the high-dimensional discretized state \cite{benner2015survey, rowley2017model}. This approach requires learning the manifold itself, characterized as a mapping between the continuous or discretized high-dimensional state and the low-dimensional latent vector, as well as a model for the dynamics within the latent space. In particular, learning a nonlinear mapping using an autoencoder or autodecoder has recently proven popular \cite{lee2020model, chen2022crom, yin2023continuous, sholokhov2023physics, he2023glasdi, wan2023evolve, serrano2023operator, raut2023arousal,otto2023learning, chen2024nonlinear}. As opposed to traditional ROMs based on linear mappings computed from methods such as the Proper Orthogonal Decomposition \cite{moore1981principal,ravindran2000reduced,holmes2012turbulence}, balanced Proper Orthogonal Decomposition \cite{willcox2002balanced,rowley2005model}, or Proper Generalized Decomposition \cite{chinesta2011overview}, the nonlinear mapping learned through an autoencoder or autodecoder enables a much more efficient compression of the high-dimensional state, especially in problems dominated by advection or exhibiting sharp gradients, for which the manifold of state trajectories is inherently nonlinear \cite{pinkus2012n,lee2020model,peherstorfer2022breaking}. As a result, data-driven ROM frameworks based on autoencoders or autodecoders have displayed the ability to forecast complex systems at a much lower cost than autoregressive operator learning methods \cite{yin2023continuous, wan2023evolve}.

However, purely data-driven approaches heavily depend on extensive training data, which can be challenging to acquire due to computational and storage limitations. For instance, the volume of training data can grow exponentially when accounting for various initial conditions, boundary conditions, and parameter combinations. Insufficient training data can result in suboptimal performance, particularly in time extrapolation problems where errors may rapidly accumulate beyond the training time horizon. A promising approach to mitigate these challenges is to combine fundamental physics principles with data-driven methods by constraining the solution space with physics-informed soft or hard constraints, thereby reducing the reliance on large datasets. Physics-informed neural networks (PINNs) \cite{dissanayake1994neural, lagaris1998artificial, raissi2019physics, cai2021physics, karniadakis2021physics, mowlavi2023optimal} exemplify this approach by incorporating PDEs into the loss function as a soft constraint. This concept has been effectively incorporated in various data-driven forecasting models, such as physics-informed CNNs \cite{zhao2023physics}, physics-informed operators learning models \cite{li2021physics, wang2021learning, goswami2023physics, navaneeth2024physics}, and physics-informed neural ODEs \cite{sholokhov2023physics}. Furthermore, fundamental physical principles like invariance \cite{ling2016machine, ling2016reynolds, xie2022data}, equivariance \cite{villar2021scalars}, symmetry \cite{gabel2023learning, choi2023symmetry}, and conservation laws \cite{hansen2023learning} are increasingly being integrated into data-driven methods. These principles enhance the robustness and applicability of the models, particularly in complex and dynamic scenarios where traditional approaches may fall short.

These advancements in physics-informed machine learning prompt a question: \textit{Can we explore more universally applicable physical principles, especially in the context of spatiotemporal surrogate modeling?} A key physical property, widely applicable to dynamical systems both with or without explicit governing equations, is the temporal smoothness of their dynamics. Thus, phenomena that evolve smoothly over time in the original state space should mirror this smooth trajectory in the reduced latent space. Autoencoders, however, always process data snapshot by snapshot independently, neglecting the temporal correlations between sequential snapshots and overlooking the fact that consecutive snapshots in latent space are likely more similar to each other than to those further away in time or from different trajectories. Additionally, the intricate structure of neural networks, combined with numerous potential local minima, often makes the learned latent vectors complex and difficult to interpret. To address these issues, recent studies have aimed at integrating smoothness into the latent space. Wan et al. \cite{wan2023evolve} implemented a cyclic consistency regularization to ensure that re-encoded reconstructed states match their original latent representations. He et al. \cite{he2023glasdi} proposed to train the autoencoder and latent dynamics model simultaneously based on predicted trajectories in both original and latent spaces. While these methods have yielded improved accuracy and efficiency, it is unclear whether their implicit approaches effectively enforce smoothness in latent space. By contrast, Desai et al. \cite{desai2024trade} proposed two explicit smoothness losses to regulate microstructure evolution in latent space, requiring the gradient of latent trajectories to maintain consistent signs and small angles. However, these conditions are quite stringent and can penalize even smooth spiraling trajectories in latent space. All these limitations highlight the challenge of finding a balance between enforcing smoothness and allowing for diverse, realistic representations.

In this study, we propose a spatiotemporal continuous operator learning framework that focuses on achieving smooth latent dynamics through jerk regularization. By directly minimizing jerk -- the derivative of acceleration -- along latent space trajectories during the training phase of the autoencoder, this approach explicitly promotes smoothness and implicitly induces sparsity of the dynamics in latent space. The incorporation of jerk regularization brings three main advantages: (1) it enhances the accuracy and speeds up the convergence of the training process for both the autoencoder and Neural Ordinary Differential Equations (neural ODEs) used to model the latent dynamics; (2) it significantly improves the smoothness of latent space trajectories; (3) it achieves increased sparsity in the latent space, promoting the discovery of intrinsic latent coordinates and eliminating the need for traditional sparsity-inducing techniques like the \(L_1\) norm. Our model also leverages recent advances in spatially and temporally continuous deep learning \cite{chen2022crom, pan2023neural, wan2023evolve}, achieving spatial continuity through the use of a conditional Implicit Neural Representation (INR)-based decoder, and temporal continuity through a neural ODE-based latent dynamics model. This comprehensive framework allows for inference at any desired spatial or temporal resolution after training, without the need for additional post-processing or super-resolution steps. We validate the efficacy of our method on a two-dimensional unsteady flow problem governed by the Navier-Stokes equations, demonstrating its potential in complex system modeling.

\section{Results}

\subsection{Spatiotemporal Continuous Operator Learning}
We consider parameterized spatiotemporal systems described by PDEs of the general form
\begin{equation}
\partial_t u(\boldsymbol{x}, t; \boldsymbol{\lambda}) + \mathcal{N}[u(\boldsymbol{x}, t; \boldsymbol{\lambda})] = 0, \quad \boldsymbol{x} \in \Omega, \quad t \in [0, T],
\label{eq:PDE}
\end{equation}
where \( u(\boldsymbol{x}, t; \boldsymbol{\lambda}) \) represents the continuous state of the system in the spatial domain \(\Omega \subseteq \mathbb{R}^d\) and temporal interval \( [0, T] \), with \(\boldsymbol{\lambda}\) as the parametrization vector and \(\mathcal{N}\) a potentially unknown nonlinear differential operator. The system is further subjected to initial and boundary conditions, denoted by \( u(\boldsymbol{x}, 0; \boldsymbol{\lambda}) = u_0(\boldsymbol{x}; \boldsymbol{\lambda}) \) and \( u(\boldsymbol{x}_b, t; \boldsymbol{\lambda}) = g(\boldsymbol{x}_b, t; \boldsymbol{\lambda}) \) for \( \boldsymbol{x}_b \in \partial\Omega \). Contrary to classical numerical methods that require complete knowledge of the governing equations, our goal is to construct a data-driven model capable of efficiently approximating PDE solutions, even without explicit governing equations. The primary goal is to leverage the intrinsic low-dimensional structure of PDE solutions for inferring the underlying dynamics from a training dataset, thereby achieving both accuracy and computational efficiency without necessitating a comprehensive understanding of the operator \(\mathcal{N}\). We assume that the training dataset consists of various solution trajectories of \eqref{eq:PDE}, where the continuous states are discretized in space using a sampling grid \( \{\boldsymbol{x}_i\}_{i=0}^N \) and in time using a fixed time interval \( \Delta t\).

\begin{figure}
    \centering
    \includegraphics[width=1.0\linewidth]{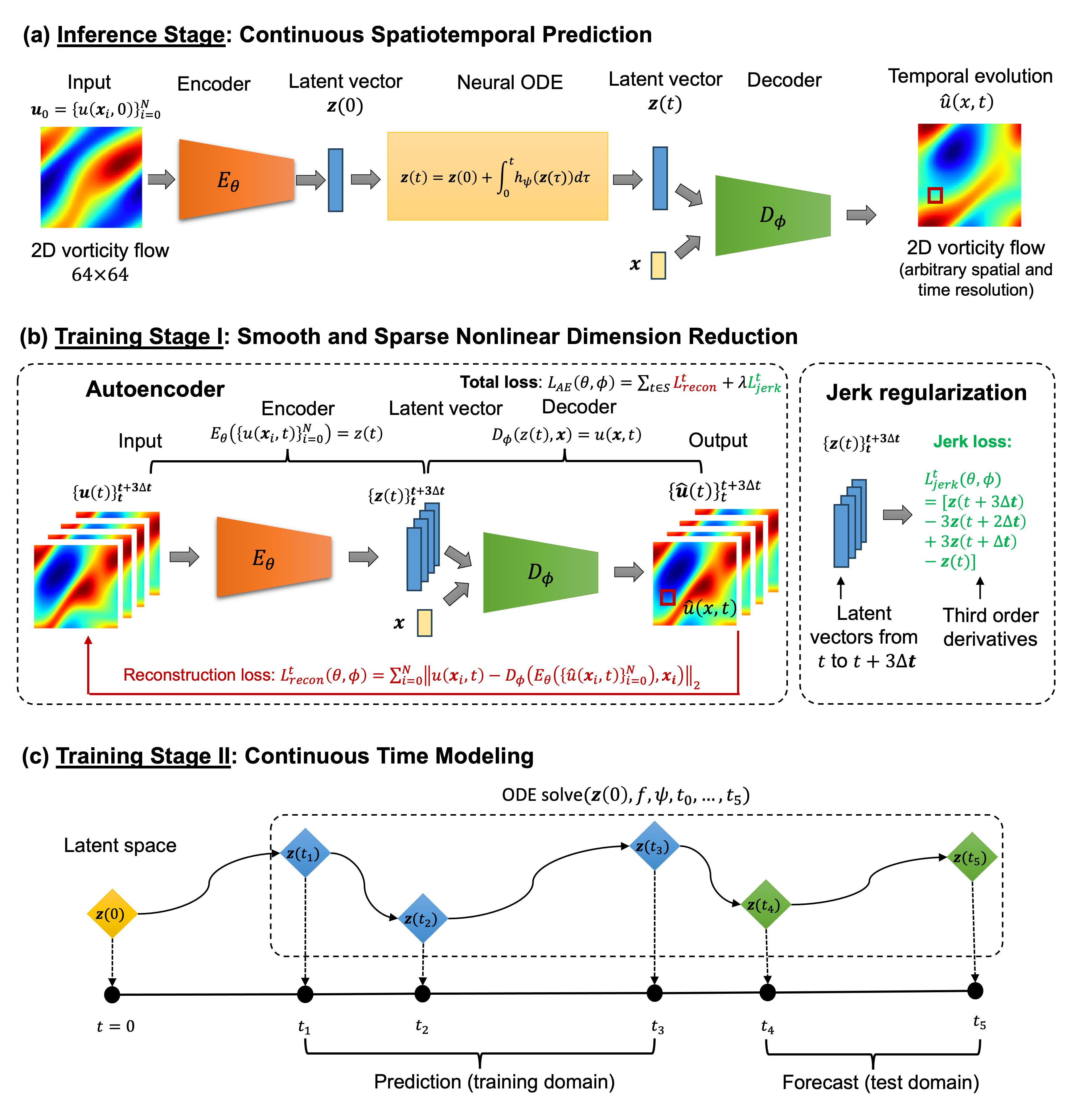}
    \caption{\textbf{Schematic of the Spatiotemporal Continuous Operator Learning Model.} \textbf{(a) Inference Stage}: The encoder transforms the initial system state \( \boldsymbol{u}(0) \) into a compressed latent representation \(\boldsymbol{z}(0)\). This initial latent vector \(\boldsymbol{z}(0)\) is then evolved using a neural ODE to obtain the corresponding latent state \(\boldsymbol{z}(t)\) at any subsequent time \(t\). A conditional INR decoder then reconstructs the continuous system state \(u(\boldsymbol{x},t)\) from \(\boldsymbol{z}(t)\) with an arbitrarily high spatial resolution by sampling at any desired spatial location. \textbf{(c) Training Stage I}: The autoencoder is trained to learn a nonlinear dimensionality reduction mapping between original and latent spaces yielding smooth and sparse latent space trajectories. To that effect, a reconstruction loss forces the output state to match the corresponding input at each time, while a jerk regularization loss enforces smoothness and sparsity by penalizing the third-order time derivative of the latent vector, approximated for a given time using four sequential latent vectors. \textbf{(c) Training Stage II}: The neural ODE is trained to learn a continuous-time model of the dynamics within the latent space. Once trained, it generates latent vectors at any desired time step, starting from the initial latent state \(\boldsymbol{z}(0)\), allowing for predictions that extend beyond the temporal scope of the training data and with variable time granularity.}
    \label{fig: schematic}
\end{figure}

Our proposed model, depicted in Figure \ref{fig: schematic}, adopts ROM-based continuous operator learning, merging nonlinear dimensionality reduction with continuous spatiotemporal modeling across space and time. During the inference stage, a CNN-based encoder transforms the initial high-dimensional system state \( \boldsymbol{u}(0)=\{u(\boldsymbol{x}_i, 0)\}_{i=0}^N \) from the ambient space of dimension \( N+1 \) into an initial latent vector \( \boldsymbol{z}(0) \) in a compressed latent space of dimension \( d_z \ll N \). Then, the low-dimensional latent vector is propagated through a neural ODE to compute a future latent vector \( \boldsymbol{z}(t) \) at an arbitrary future time \( t \). Propagating the latent vector instead of the high-dimensional state significantly accelerates the inference because the latent dimension \( d_z \) is significantly smaller than the ambient space dimension \( N+1 \). Subsequently, a conditional INR-based decoder maps the future latent vector \( \boldsymbol{z}(t) \) back to the predicted continuous state \( \hat{\boldsymbol{u}}(\boldsymbol{x}, t) \), enabling predictions at arbitrary spatial and temporal resolutions.

The training process is split into two stages. During stage I, illustrated in Figure \ref{fig: schematic} (b), we train an autoencoder, consisting of an encoder \( E_\theta \) with parameters \( \theta \) and a decoder \( D_\phi \) with parameters \( \phi \), to learn an invertible nonlinear map between the ambient space and the compressed latent space that yields smooth and sparse solution trajectories in latent space. We employ a standard reconstruction loss \(L_{recon}^t \), whose details are described in Section \ref{method:reconstruction loss}, to ensure that the continuous states recovered from the decoder are similar to the high-dimensional states given to the encoder. Crucially, we also ensure that trajectories in latent space are smooth by including a jerk regularization loss that penalizes the third-order time derivative of the latent vector at each time \( t \), which encourages changes in acceleration along the latent trajectory to be small without restricting the magnitude and direction of the velocity and acceleration. Using a finite difference approximation of the acceleration, the jerk regularization loss \( L_{jerk}^t \) at time \( t \) is calculated as
\begin{equation}
    L_{jerk}^t = \| \boldsymbol{z}(t+3\Delta t) - 3\boldsymbol{z}(t+2\Delta t) + 3\boldsymbol{z}(t+\Delta t) - \boldsymbol{z}(t) \|_2^2,
\end{equation}
where \( \boldsymbol{z}(t) \), \( \boldsymbol{z}(t+\Delta t) \), \( \boldsymbol{z}(t+2\Delta t) \), and \( \boldsymbol{z}(t+3\Delta t) \) are the latent vectors obtained by passing through the encoder a sequence of high-dimensional states \( \boldsymbol{u}(t) \), \( \boldsymbol{u}(t+\Delta t) \), \( \boldsymbol{u}(t+2\Delta t) \), and \( \boldsymbol{u}(t+3\Delta t) \) belonging to the same solution trajectory, and \( \|\cdot\|_2 \) denotes the \(L_2\) norm of the vector. The total loss at time \( t \) for the training of the autoencoder is a weighted combination of the reconstruction loss \(L_{recon}^t \) and the jerk regularization loss \( L_{jerk}^t \). The complete training procedure for the autoencoder is described in Section \ref{method:stage1}, and a more detailed description of the jerk regularization is provided in Section \ref{method:jerk}.

Once the autoencoder is trained, we convert snapshots across all trajectories in the training dataset from the ambient space into their latent space representations. This transformed latent dataset then serves as the basis for training the neural ODE during the training stage II, which is shown in Figure \ref{fig: schematic} (c). The neural ODE with parameters \( \psi\) learns the temporal evolution of the latent vectors, allowing the neural network to capture complex system dynamics within a condensed framework. One notable advantage of this method is its flexibility for non-uniform temporal sampling in the inference stage, offering the ability to interpolate latent states at arbitrary time steps once the model is trained. More details on the training of the neural ODE are given in Section \ref{method:stage2}.

Our approach contrasts with traditional autoencoders, which are limited by the need for fixed spatial resolutions in the output. Instead, following other recent works \cite{chen2022crom, pan2023neural, wan2023evolve, yin2023continuous}, we use a conditional INR for the decoder, capable of reconstructing arbitrary high-dimensional states \( \{u(\boldsymbol{x}_i, t)\}_{i=0}^M \) from training datasets that may be of lower resolution \( \{u(\boldsymbol{x}_i, t)\}_{i=0}^N \), where \( M \) is independent of \( N \) and can be higher or lower than \( N \). By conditioning the INR on the latent vector \( z(t) \), the decoder is capable of reconstructing not just a static scene, but a sequence of evolving system states over time. The details of the conditional INR model are described in Section \ref{method:inference}.

\section{Numerical results: Navier-Stokes Equations}

\subsection{Dataset Preparation}

In this section, we demonstrate the effectiveness of the proposed framework using the Navier-Stokes equations, which govern the motion of viscous fluids. Being able to model and predict the evolution of fluid flows accurately has implications ranging from improving the design of aerodynamic vehicles to enhancing weather prediction accuracy. Specifically, we demonstrate our method on a 2D problem governed by the incompressible Navier-Stokes equations in vorticity form with an added force term:

\begin{align}
    \frac{\partial w}{\partial t} + u \cdot \nabla w &= \nu \Delta w + f, \\
    \nabla \cdot u &= 0
\end{align}

where \( w \) represents the vorticity, \( u \) is the velocity field, \( \nu \) is the fluid viscosity set at \( 1 \times 10^{-3} \), and \( f \) is the forcing function given by \( f(x_1, x_2) = 0.1(\sin(2\pi(x_1 + x_2)) + \cos(2\pi(x_1 + x_2))) \). The simulation is conducted on the 2D spatial domain \([0, 1] \times [0, 1]\), and the initial condition for the vorticity \( w_0(x) \) is sampled from a distribution \( \mu = N(0, 7^{3/2} (-\Delta + 49I)^{-2.5}) \) subject to periodic boundary conditions. We use the dataset of solution trajectories from \cite{li2020fourier}.

The original resolution of the simulation data was \(256 \times 256\), which was later downsampled to \(64 \times 64\) for training purposes. The total dataset consists of 50 time steps with \( \Delta t = 1 \). The initial 10 time steps were removed due to inherent noise and diminished relevance to the underlying dynamics. The data from time \( t=11 \) to \( t=40 \) is designated for training, while data from \( t=41 \) to \( t=50 \) is set aside for time extrapolation testing. Of the 1000 total trajectories, 900 trajectories were used for training, and the remaining 100 trajectories for testing.

\subsection{Comparative Analysis of Model Accuracy}

\begin{figure}
  \centering
  \includegraphics[width=1.0\linewidth]{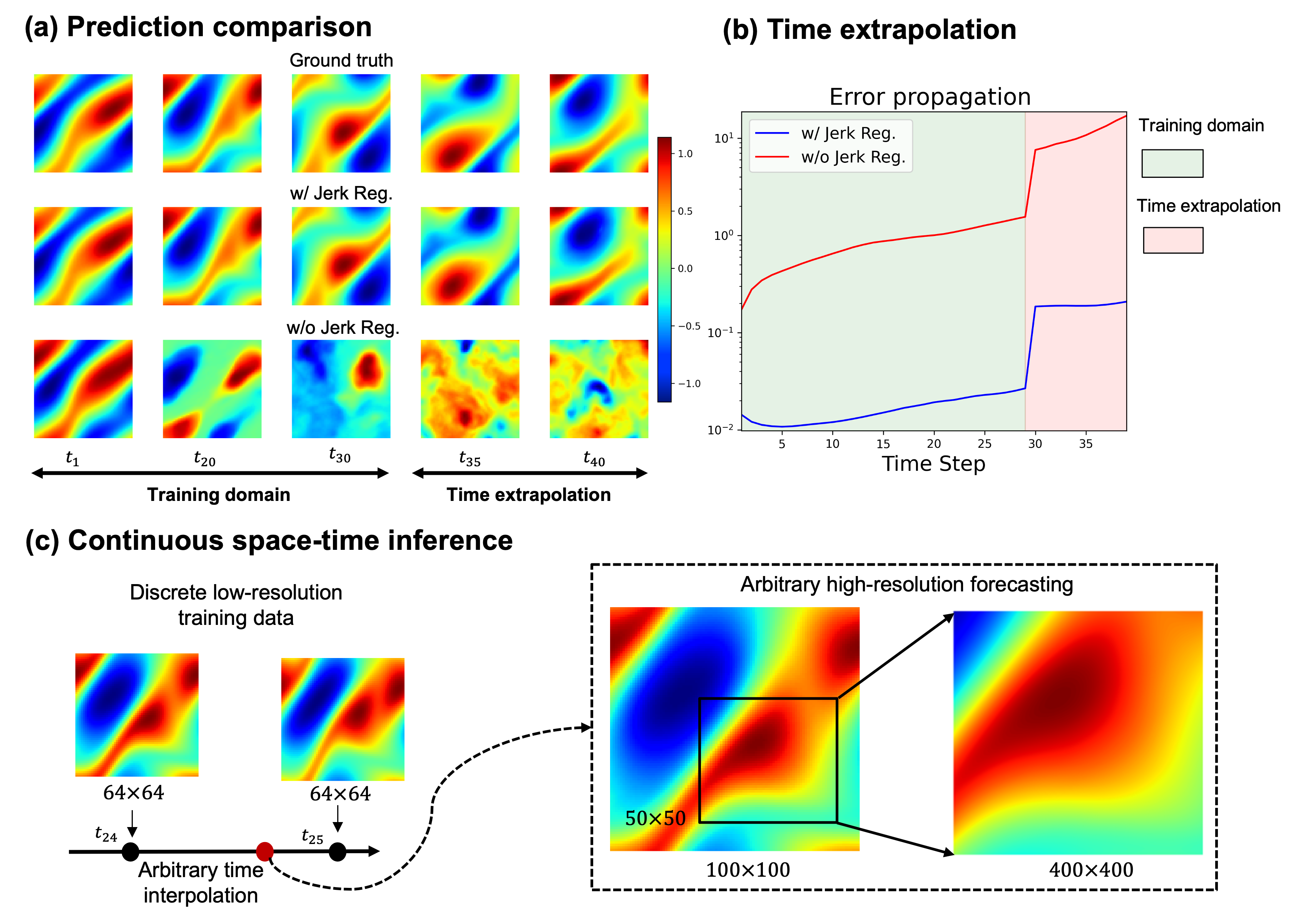}
  \caption{\textbf{Comparison of ground truth vorticity fields and model predictions for the Navier-Stokes equations example.} \textbf{(a)} Time interpolation and extrapolation comparisons in the test set. \textbf{(b)} Error propagation inside (green area) and outside (pink area) the training time domain. \textbf{(c)} The proposed method can conduct arbitrary high-resolution forecasting in space and time.}
  \label{fig: fig_NS_compare}
\end{figure}

Figure \ref{fig: fig_NS_compare} illustrates the enhanced predictive capabilities of our model when incorporating jerk regularization. Panel (a) of the figure compares the predictions of the model with and without jerk regularization in the test set. It is evident that jerk regularization markedly improves model predictions within the training time domain. Remarkably, this improvement extends to the time extrapolation range, where the model with jerk regularization maintains robust predictions. Panel (b) of Figure \ref{fig: fig_NS_compare} presents the average Relative Mean Square Error (RMSE) \cite{li2020fourier} across the dataset inside and outside the training time domain. Here, the green shaded area represents the training time domain, while the red shaded area indicates the time extrapolation range. The model without jerk regularization exhibits a high initial error, which tends to accumulate more rapidly compared to the model with jerk regularization. Notably, in the time extrapolation range, the RMSE of the non-regularized model increases quickly, whereas the RMSE of the jerk-regularized model remains relatively low and shows only a slight increase.

Panel (c) of Figure \ref{fig: fig_NS_compare} demonstrates the super-resolution capabilities of our proposed method, using training data of low resolution (\( 64 \times 64 \)). Conventional data-driven models like U-Net \cite{gupta2022towards, mialon2023self}, ConvLSTM \cite{ren2022phycrnet,mavi2023unsupervised, rao2023encoding}, and FNO \cite{li2020fourier} are not capable of predicting solutions at higher spatial resolutions than that of their training data without post-processing techniques. These models either necessitate paired low and high-resolution training data (as in the case of U-Net and ConvLSTM) or require high-resolution initial conditions during the inference stage to produce high-resolution predictions, as with FNO. Moreover, these methods are also incapable of predicting solutions at finer time intervals than that of their training data. In contrast, {\it our model is trained solely on low-resolution data yet is capable of interpolating at any spatial and temporal resolution}. As depicted in panel (c), our method allows for inference between time steps \(t_{24}\) and \(t_{25}\), where no training data exists. This feature enables detailed examination of specific areas, offering high-resolution solutions through further inference, underscoring the versatility and effectiveness of our proposed approach.

\subsection{Comparative Analysis of Smoothness and Sparsity}

\begin{figure}[ht]
  \centering
  \includegraphics[width=1.0\linewidth]{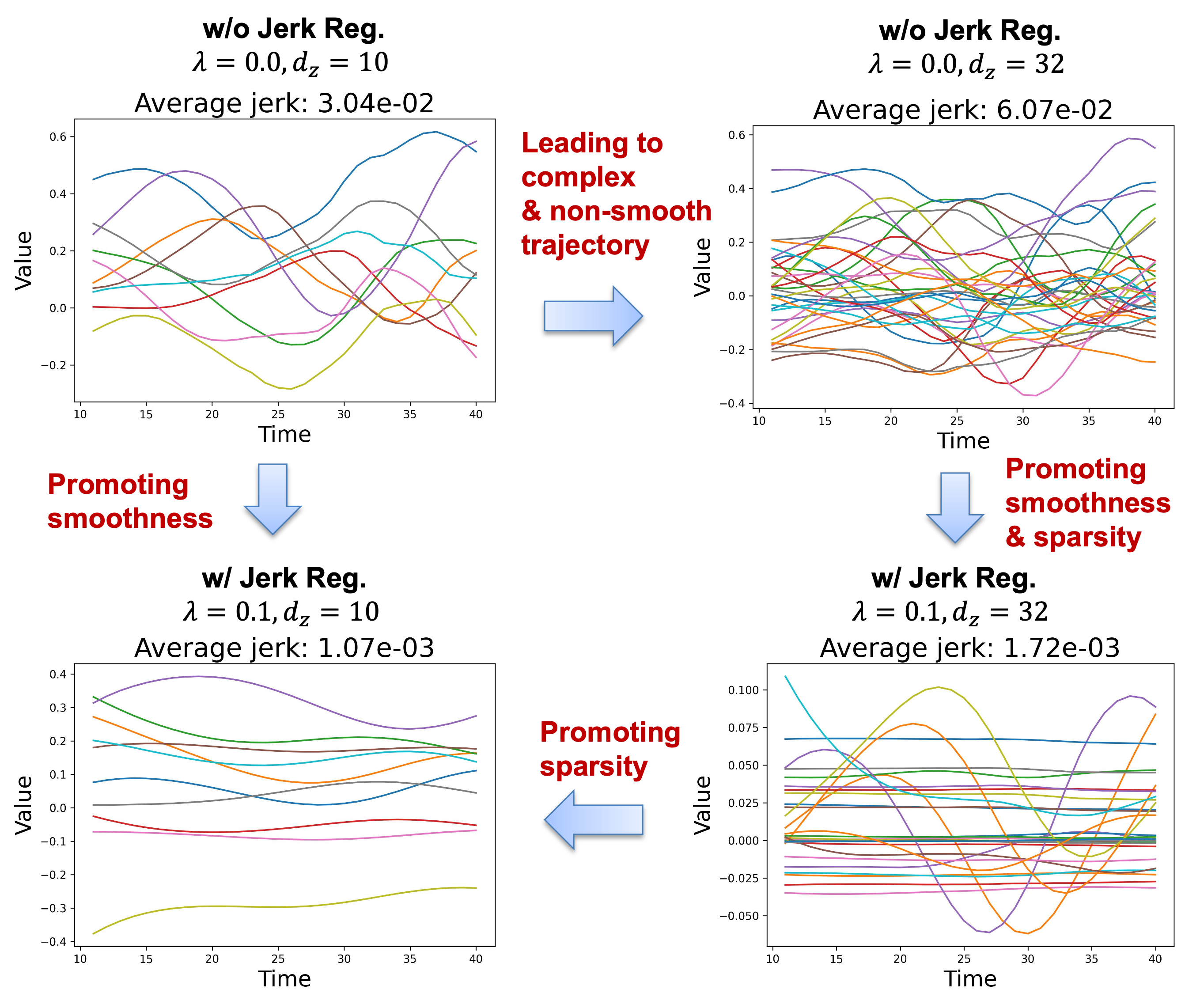}
  \caption{\textbf{Jerk regularization promotes smoothness and sparsity in latent space.} The average jerk is calculated by taking the mean value of the jerk measurements across all snapshots in a trajectory.}
  \label{fig: fig_NS_AE_jerk}
\end{figure}

Figure \ref{fig: fig_NS_AE_jerk} elucidates the effect of jerk regularization on the latent dynamics across varying latent space dimensions. The top left panel depicts a latent space trajectory in the test set using a latent space dimension of \(d_z=10\), obtained from an autoencoder trained without jerk regularization. The trajectory is visualized in terms of the time series of the various components of the latent vector, and the average jerk metric is given by the same formula as the jerk regularization. Despite a generally smooth profile, each latent coordinate contains segments that exhibit roughness, complicating the training of the neural ODE. Increasing the latent space dimension to \(d_z=32\), as shown in the top right panel, results in a significantly more complex trajectory with all latent coordinates active and a nearly twofold increase in the average jerk metric from \(3.04 \times 10^{-2}\) to \(6.07 \times 10^{-2}\), presenting additional challenges in learning the latent dynamics.

The lower panels of Figure \ref{fig: fig_NS_AE_jerk} demonstrate the ability of jerk regularization to enhance smoothness and promote sparsity. For the lower-dimensional case (\(d_z=10\)), as seen in the bottom left panel, including jerk regularization when training the autoencoder yields a markedly smoother trajectory and a notable 27-fold reduction in the average jerk metric, which facilitates learning the latent dynamics. In the high-dimensional case (\(d_z=32\)), presented in the bottom right panel, including jerk regularization leads to a markedly smoother and simpler-looking latent space trajectory, with a 35-fold drop in the average jerk metric. Interestingly, {\it besides enhancing smoothness, jerk regularization also encourages sparsity of the latent vector, with most latent coordinates remaining constant over time and only ten of them exhibiting variability}. This observation suggests that training an autoencoder with jerk regularization can guide the selection of an optimal latent space dimension in a data-driven fashion, providing insights into the number of intrinsic latent coordinates typically deduced for various systems either using physical knowledge or through a trial-and-error approach.

\subsection{Identifying Intrinsic Coordinates}

To investigate the efficacy of jerk regularization in identifying intrinsic coordinates \cite{champion2019data}, we conducted a systematic comparison of models with varying latent space dimensions, both with and without jerk regularization. The findings are shown in Table \ref{tab:table_comparison_NS}. We determined the number of active coordinates for each model by analyzing the variance of each latent element against a predefined threshold. Elements with variances below this threshold were classified as inactive and thus considered redundant. It also needs to be noted that the threshold is a hyperparameter that may need adjustment depending on the specific context. Setting the threshold excessively high could lead to an underestimation of the number of active coordinates, while a threshold that is too low might result in an overestimation. To determine an appropriate threshold, we initially examine a selection of random latent trajectories visually. For our analysis, we adjusted the thresholds for different latent space dimensions, namely 8, 10, 16, 32, and 64, to values of \(1 \times 10^{-4}\), \(5 \times 10^{-5}\), \(1 \times 10^{-4}\), \(1 \times 10^{-5}\), \(1 \times 10^{-5}\), and \(5 \times 10^{-4}\) respectively.

The results, presented in Table \ref{tab:table_comparison_NS}, reveal that models trained without jerk regularization utilize all latent coordinates, which correlates with higher reconstruction (stage I) and prediction (stage II) errors, including Mean Squared Error (MSE) and RMSE values. Conversely, the incorporation of jerk regularization not only effectively decreases the number of active coordinates but also markedly reduces the MSE and RMSE for both reconstruction and prediction. Despite the latent space dimension ranging from 8 to 64, our findings indicate that a consistently low number of 8 to 10 latent coordinates are active, which may correspond to the number of intrinsic coordinates describing the dynamics of the system.

When comparing stage II MSE and RMSE across various latent space dimensions, the results suggest that a smaller latent space dimension often leads to superior model performance. Indeed, increasing the latent dimension from 10 to 64 results in a substantial rise in both MSE and RMSE in stage II. This highlights the importance of being able to identify the number of intrinsic latent coordinates so that the right latent space dimension for optimal model performance can be selected.

\begin{table}[ht]
\centering
\small
\setlength{\tabcolsep}{3.5pt} 
\renewcommand{\arraystretch}{1.5} 
\begin{tabular}{c|cc|cc|cc|cc|cc}
\hline
\textbf{Latent dims.} & \multicolumn{4}{c|}{\textbf{w/o Jerk Reg. ($\lambda=0.0$)}} & \multicolumn{6}{c}{\textbf{w/ Jerk Reg. ($\lambda=0.1$)}} \\
\cline{2-11} 
 & Jerk & Active & MSE & RMSE & Jerk & Active & MSE & RMSE & MSE & RMSE \\
 & Loss & Coord. & -Stage I & -Stage I & Loss & Coord. & -Stage I & -Stage I & -Stage II & -Stage II \\
\hline
8 & 4.60E-02 & 8.00 & 6.16E-03 & 6.30E-02 & 1.46E-03 & 7.87 & 1.23E-04 & 1.02E-02 & 5.21E-03 & 3.89E-02 \\
\underline{\textbf{10}} & 8.12E+00 & 10.00 & 3.00E-03 & 2.52E-02 & \underline{\textbf{1.37E-03}} & \underline{\textbf{9.10}} & \underline{\textbf{6.62E-05}} & \underline{\textbf{8.47E-03}} & \underline{\textbf{3.45E-04}} & \underline{\textbf{1.78E-02}} \\
16 & 2.01E-01 & 15.99 & 1.28E-02 & 8.75E-02 & 1.70E-03 & 9.57 & 1.04E-04 & 1.02E-02 & 3.70E-04 & 1.88E-02 \\
32 & 6.76E-02 & 32.00 & 5.24E-03 & 5.25E-02 & 2.49E-03 & 9.55 & 4.84E-04 & 1.13E-02 & 2.52E-01 & 5.27E-01 \\
64 & 1.78E+01 & 64.00 & 1.62E-02 & 8.47E-02 & 1.68E-03 & 8.35 & 1.72E-04 & 8.57E-03 & 2.53E-01 & 5.29E-01 \\
\hline
\end{tabular}
\caption{\textbf{Comparison of latent space dimensions and active coordinates in the test set.} Stage I error refers to the reconstruction error of the autoencoder, while stage II error refers to the prediction error of the entire operator learning model. These errors correspond to the two training stages detailed in Section \ref{sec:two-stage}.}
\label{tab:table_comparison_NS}
\end{table}

\subsection{Comparative Analysis of Convergence Rate}

\begin{figure}
  \centering
  \includegraphics[width=1.0\linewidth]{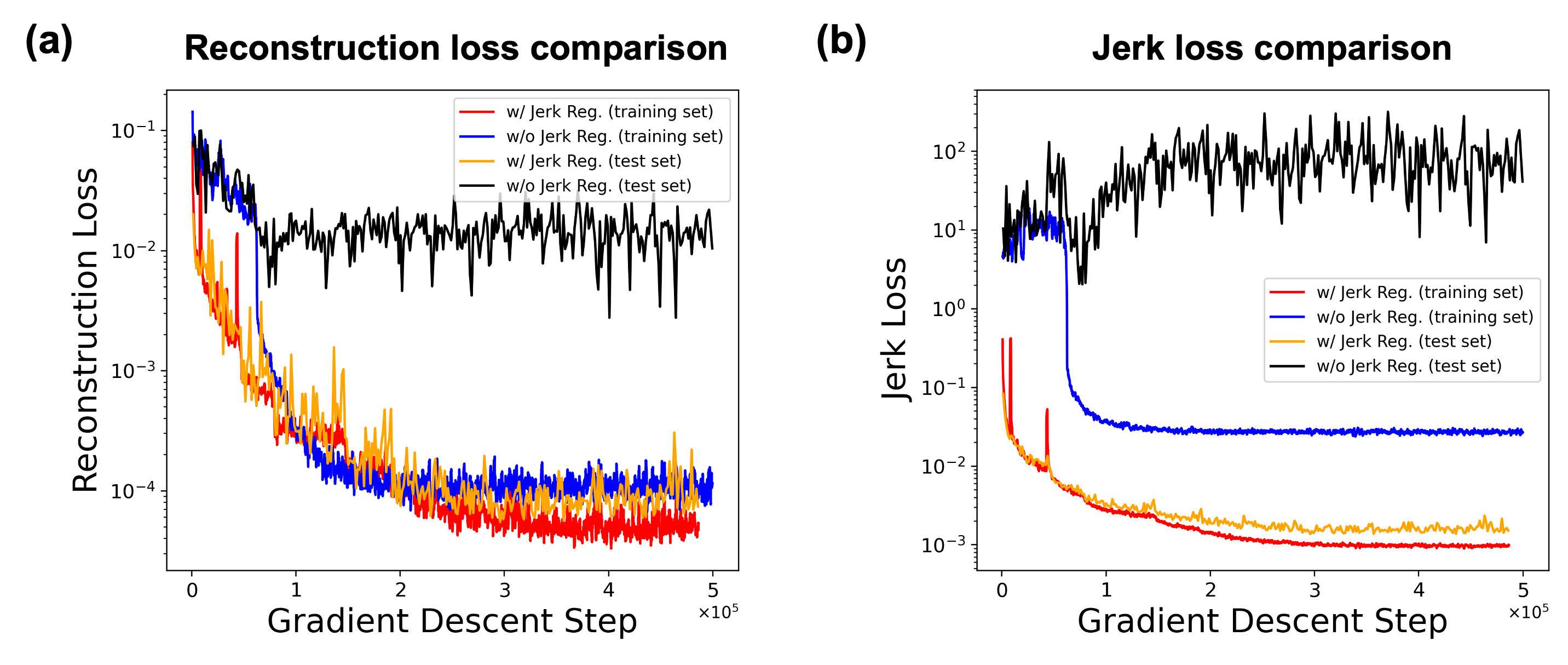}
  \caption{\textbf{Comparison of the stage I loss history with and without jerk regularization for a latent space dimension of 10.} \textbf{(a)} MSE loss comparison for training and test sets. \textbf{(b)} Jerk loss comparison for training and test sets.}
  \label{fig: fig_NS_AE_loss}
\end{figure}

In this section, we analyze the impact of jerk regularization on the loss history during training stage I of models with and without jerk regularization, as visualized in Figure \ref{fig: fig_NS_AE_loss}. The comparison between the red and blue curves in panel (a) indicates that jerk regularization considerably accelerates convergence within the initial 20\% of gradient descent iterations. 

Furthermore, it is evident that jerk regularization substantially mitigates overfitting. Indeed, while the final training reconstruction loss of the model without jerk regularization is only slightly larger than that of the model with jerk regularization, this difference is greatly amplified on the test set. Specifically, the model lacking jerk regularization exhibits a final test MSE loss around a hundred times greater than that of the model with jerk regularization, whose test MSE loss remains comparable to its training MSE loss. 

Clearly, jerk regularization helps the training stage I find a much better local minimum in the parameter landscape of the autoencoder. To understand this property, the elevated test MSE loss of the non-regularized model is further elucidated in panel (b) of Figure \ref{fig: fig_NS_AE_loss}. We observe that the test jerk loss without jerk regularization is more than four orders of magnitude higher than with jerk regularization, implying that the autoencoder without jerk regularization did not learn the true manifold of solution trajectories, leading to complex and erratic latent dynamics. Conversely, thanks to the enforcement of the smoothness property of the system dynamics, the model with jerk regularization exhibits low jerk loss even in the test set. Taken together with the low test MSE loss of the regularized model, we conclude that jerk regularization helps the autoencoder in the training stage I discover the true manifold of solution trajectories.

\subsection{Comparative Analysis of Jerk Coefficient}

In the proposed method for training the autoencoder, the overall loss function is composed of two main components: the reconstruction loss and the jerk loss. The latter is modulated by a specific parameter called jerk coefficient \( \lambda \). This coefficient plays a crucial role in determining the influence of the jerk loss on the total loss. Adjusting the value of \( \lambda \) allows for fine-tuning the balance between the reconstruction accuracy and the smoothness of the trajectory. Therefore, selecting an appropriate value for the jerk coefficient is essential for the effectiveness of the training process.

\begin{figure}
  \centering
  \includegraphics[width=0.6\linewidth]{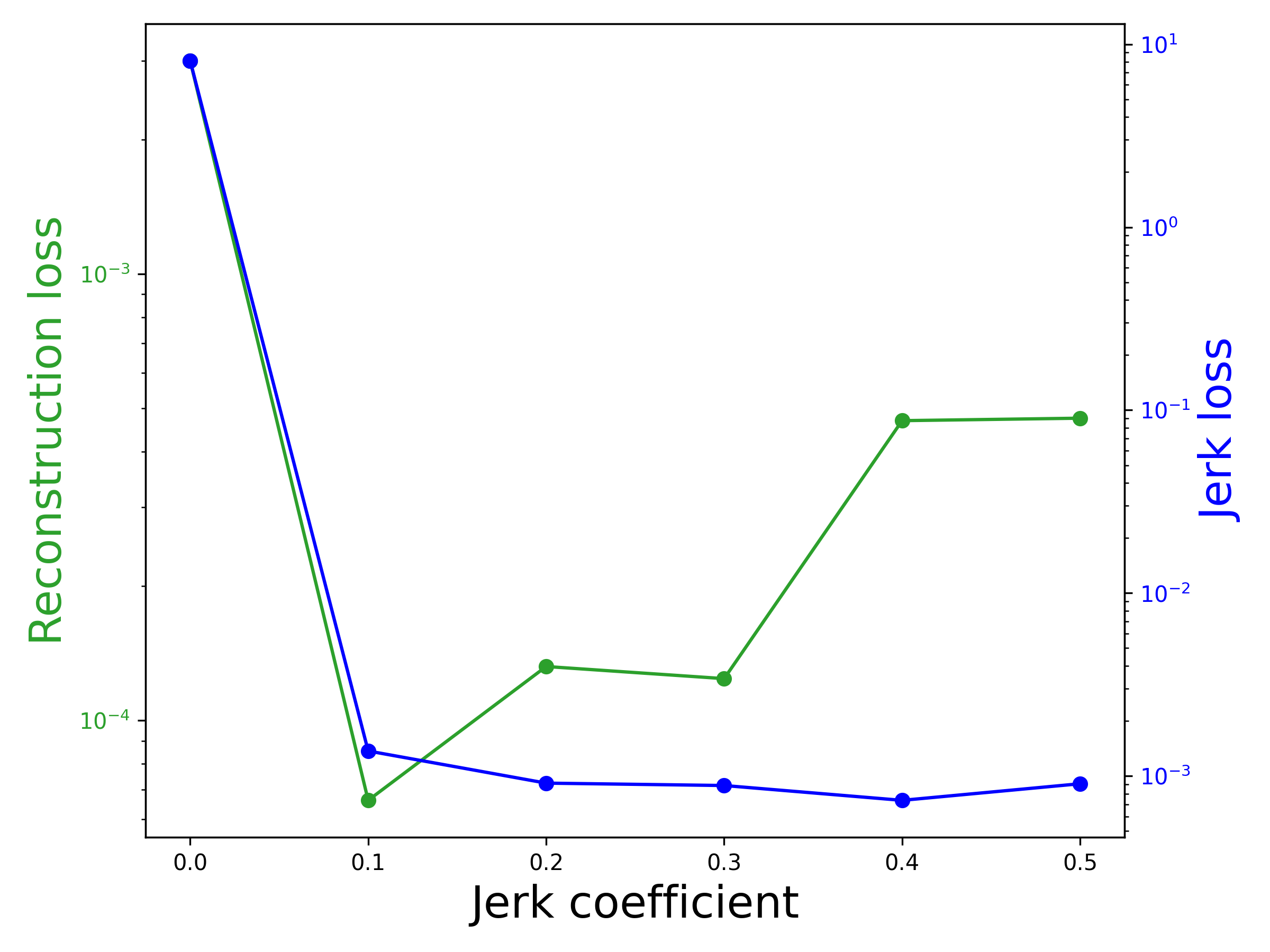}
  \caption{\textbf{Comparison of different jerk coefficients for autoencoder in the test set.}}
  \label{fig: fig_AE_compare_latent_coef}
\end{figure}

Figure \ref{fig: fig_AE_compare_latent_coef} illustrates the variation in reconstruction loss and jerk loss across a range of jerk coefficients ranging from 0.0 to 0.5. It is observed that a jerk coefficient of 0.1 yields the lowest MSE, alongside a low jerk loss. As the jerk coefficient increases beyond 0.1, the MSE starts to rise, indicating the disproportionate influence of jerk regularization on the total loss. This suggests that an optimal balance between reconstruction loss and jerk loss is essential. Within the context of the Navier-Stokes equations addressed in this study, the optimal jerk coefficient is determined to be 0.1, striking a balance that minimizes MSE while maintaining low jerk loss.

\section{Methods}

\subsection{Encoder}
The goal of the encoder is to efficiently reduce the continuous state at a given time into a more compact, lower-dimensional representation, ensuring the preservation of crucial dynamic characteristics. For a given time instance $t$, the continuous state of the system, denoted as ${u(\boldsymbol{x}, t)}$, is discretized in space using a sampling grid $\{\boldsymbol{x}_i\}_{i=0}^N$. The resulting high-dimensional state, expressed as $\boldsymbol{u}_t={\{u(\boldsymbol{x}_i, t)}\}_{i=0}^N$, constitutes the input to the encoder, whose objective is to distill $\boldsymbol{u}_t$ into key features within a latent space. Given the high-dimensional state $\boldsymbol{u}_t$, a compressed latent representation $\boldsymbol{z}(t)$ can be obtained as
\begin{equation}
    \boldsymbol{z}(t) = E_\theta({\{u(\boldsymbol{x}_i, t)}\}_{i=0}^N),
\end{equation}
where $E_\theta$ symbolizes the encoder's operation, parameterized by the neural network parameters $\theta$.

The encoding process employs a CNN based on the ResNet architecture \cite{he2016deep}, specifically utilizing ResNet50 as the foundational model. We have adapted this model to effectively process Navier-Stokes equations datasets. To accommodate our dataset's nature, the initial channel number is set as one, targeting only the vorticity field. After processing through four ResNet blocks, a fully connected layer then reduces the latent dimension from 2048 to the final latent dimension $d_z$, such as 10 or 16. To stabilize the learning process in the encoder, we employed Kaiming normal initialization \cite{he2015delving}, ensuring consistent variance in the network's output across layers, particularly before and after ReLU activation functions.

\subsection{Latent Dynamics: Neural ODE}

Neural ODEs \cite{chen2018neural} offer a flexible method to parameterize the time derivative of a latent state varying continuously in time using a neural network. In our work, we use a neural ODE to model the continuous-time dynamics of the latent vector \( \boldsymbol{z}(t) \), which can be expressed as:
\begin{equation}
    \frac{d\boldsymbol{z}(t)}{dt} = h_\psi(\boldsymbol{\boldsymbol{z}}(t)),
\end{equation}
where \( h_\psi(\cdot) \) is a neural network with parameters \( \psi \) that models the rate of change of \( \boldsymbol{z} \). Due to the universal approximation property of neural networks \cite{hornik1989multilayer}, this formulation allows for an arbitrary complex behavior in the latent space, with \( h_\psi(\cdot) \) capturing the system's dynamics.

To compute the trajectory of \( \boldsymbol{z}(t) \) over an arbitrary time interval, one can employ an ODE solver, taking \( \boldsymbol{z}(t_0) \) as the initial condition and integrating forward in time, which can be represented as:
\begin{equation}
    \boldsymbol{z}(t) = \boldsymbol{z}(0) + \int_0^t h_\psi(\boldsymbol{z}(\tau))d\tau.
\end{equation}
Using an ODE solver with adaptive step size, this continuous-time representation offers the advantage of being able to evaluate \( \boldsymbol{z}(t) \) at any desired point in time, without being constrained by fixed time steps as in traditional recurrent neural networks or autoregressive time-stepping methods used by DeepONet and FNO. Further, this formulation lends itself to well-established techniques for stability and error control in numerical solutions of ODEs \cite{chen2018neural}. After tuning with different hyperparameters, such as learning rate, batch size, and neural network structures, we apply a Multi-Layer Perceptron (MLP) with 5 hidden layers for \( h_\psi(\cdot) \). Each of these layers comprises 512 neurons, and they are each combined with SiLu activation functions \cite{elfwing2018sigmoid}.

\subsection{Decoder: Conditional INR}
The goal of the decoder is to map the compressed latent state at a given time back to the corresponding continuous system state. Contrary to classical autoencoder approaches where the decoder returns a discretized high-dimensional state, we formulate the decoder using an Implicit Neural Representation (INR) to preserve the spatial continuity of the reconstructed state. INRs are a method to model continuous fields as functions using neural networks, enabling compact and smooth representations of complex data like 2D images or 3D shapes \cite{park2019deepsdf, sitzmann2020implicit}. A vanilla INR only takes in coordinates as input and can only be used to represent a single image or shape. To enable the same INR neural network to enable the reconstruction of continuous system states at various times, each corresponding to a given latent vector, we condition the INR on the latent vector \cite{park2019deepsdf, chen2022crom}. Given the latent vector \( \boldsymbol{z}(t) \) and spatial coordinate \( \boldsymbol{x} \), the continuous state \( u(\boldsymbol{x}, t) \) can be reconstructed as
\begin{equation}
    u(\boldsymbol{x}, t) = D_\phi(\boldsymbol{z}(t), \boldsymbol{x}) ,
\end{equation}
where \( D_\phi \) denotes the decoder operation, parameterized by \( \phi \). Here, the latent vector \( \boldsymbol{z}(t) \) serves as the condition, and the spatial coordinate \( \boldsymbol{x} \) acts as a query. After tuning hyperparameters and neural networks, a 7-hidden layers MLP is used for the conditional INR. Each layer has 512 neurons and a SiLU activation.

\subsection{Reconstruction Loss}
\label{method:reconstruction loss}

To train the autoencoder, we employ a reconstruction loss which quantifies the discrepancy between the original and reconstructed states. In mathematical terms, the reconstruction loss \( L_{recon}^t \) for each time step is formulated as
\begin{equation}
    L_{recon}^t = \frac{1}{N} \sum_{i=0}^{N} \| u(\boldsymbol{x}_i, t) - D_\phi(E_\theta({u(\boldsymbol{x}_j, t)}\}_{j=0}^N), \boldsymbol{x}_i) \|_2^2.
\end{equation}

Given a batch of time steps, the overall reconstruction loss \( L_{recon} \) is computed by summing the losses across all time steps in the batch:
\begin{equation}
    \label{equ:rec}
    L_{recon} = \frac{1}{4|S|} \sum_{t \in S} L_{recon}^t = \frac{1}{4|S|} \frac{1}{N}\sum_{t \in S} \left[ \sum_{i=0}^{N} \| u(\boldsymbol{x}_i, t) - D_\phi(E_\theta({u(\boldsymbol{x}_j, t)}\}_{j=0}^N), \boldsymbol{x}_i) \|_2^2 \right],
\end{equation}
where \( S \) represents the set of randomly chosen time steps in a batch from different trajectories. For each time step, we use four consecutive snapshots in time. This loss ensures that the autoencoder learns an invertible mapping between the continuous state $u(\boldsymbol{x}, t)$ and the corresponding latent vector $\boldsymbol{z}(t)$. 

\subsection{Jerk Regularization}
\label{method:jerk}

Jerk, a term primarily rooted in classical mechanics, represents the third temporal derivative of a position function or the rate of change of acceleration. Historically described for motion trajectories, its mathematical representation for a position function \( s(t) \) is
\begin{equation}
    J(t) = \frac{d^3s(t)}{dt^3}.
\end{equation}

Within the scope of our continuous deep learning model, jerk serves as a metric gauging the rate at which the dynamics of the latent vector undergo sudden changes. By integrating jerk regularization, we essentially append a higher-order regularization constraint on solution trajectories within the latent space. The jerk loss for a given time $t$ is evaluated through a forward finite-difference approximation of the continuous jerk:
\begin{equation}
    L_{jerk}^t = \frac{1}{d_z} \| \frac{d^3\boldsymbol{z}(t)}{dt^3} \|_2 \propto \frac{1}{d_z} \| \boldsymbol{z}(t+3\Delta t) - 3\boldsymbol{z}(t+2\Delta t) + 3\boldsymbol{z}(t+\Delta t) - \boldsymbol{z}(t) \|_2^2.
\end{equation}

Introducing jerk regularization into our framework has two core motivations:

(1). \textbf{Smoothness}: A core principle of minimizing jerk is the procurement of trajectories in the latent space that are devoid of abrupt transitions, thereby guaranteeing that the autoencoder model preserves the smoothness of trajectories in the original state space. By ensuring the smoothness of trajectories in the latent space, the jerk regularization ensures that the autoencoder learns the most accurate low-dimensional manifold containing solution trajectories of the original system.

(2). \textbf{Speeding up training}: Latent trajectories exhibiting higher smoothness intrinsically alleviate the training intricacies of neural ODEs. By precluding sudden shifts and rapid accelerations in the latent vector sequences, the latent space dynamics are simpler and therefore easier to learn. This naturally helps the training of the neural ODE model, which can more rapidly converge to the latent space dynamics.

In a given batch of randomly selected time steps, the total jerk regularization loss \( L_{jerk} \) is calculated as
\begin{equation}
    \label{equ:jerk}
    L_{jerk} = \frac{1}{|S|} \sum_{t \in S} L_{jerk}^t = \frac{1}{|S|} \frac{1}{d_z} \sum_{t \in S} \| \boldsymbol{z}(t+3\Delta t) - 3\boldsymbol{z}(t+2\Delta t) + 3\boldsymbol{z}(t+\Delta t) - \boldsymbol{z}(t) \|_2^2,
\end{equation}
where the latent vectors are obtained by passing four consecutive states belonging to the same trajectory to the encoder. Based on the batch size, the total jerk regularization loss will time the batch size. Eq.~\ref{equ:jerk} accumulates the jerk regularization loss across all time steps within the batch. Evaluating the jerk regularization loss over multiple batches penalizes jerk effects over entire solution trajectories of the latent vector.

Crucially, it was observed that the jerk regularization, beyond ensuring smoothness, naturally fostered sparsity in the latent vector's components, with only a subset of them being active over the solution trajectory. This effect is noteworthy as sparsity often results in more interpretable and efficient representations. Although no explicit \(L_1\) or \(L_2\) norm-based regularization was applied, the resulting latent vectors exhibited this desirable property, underscoring the multi-faceted advantages of jerk regularization.

In essence, the integration of jerk regularization not only enables the model to preserve smoothness and promote sparsity of solution trajectories in the latent space but also proves pivotal in improving the efficiency of the subsequent neural ODE training phase.

\subsection{Two-Stage Training Scheme}
\label{sec:two-stage}

In pursuit of a cohesive model that seamlessly blends nonlinear dimension reduction and continuous latent dynamics learning, we employ a two-stage training scheme.

\subsubsection{Stage I: Autoencoder Training}
\label{method:stage1}

Initially, a comprehensive dataset of training trajectories corresponding to various initial conditions is collected, each containing a series of snapshots spanning from \(t=0\) to \(t=T\). During the first training stage, the autoencoder learns a nonlinear dimensionality reduction mapping between the original and latent spaces. To this effect, the encoder and decoder are simultaneously optimized using a dual-objective loss function including both the reconstruction loss (section \ref{method:reconstruction loss}) and the jerk regularization loss (section \ref{method:jerk}). Mathematically, the loss function for this stage is represented as
\begin{equation}
    \label{equ:stageIloss}
    L_{AE} = L_{recon} + \lambda L_{jerk},
\end{equation}
where \(L_{recon}\) and \(L_{jerk}\) are respectively the reconstruction and jerk regularization losses defined previously, and jerk coefficient \(\lambda\) is a hyperparameter balancing the two losses.

To implement mini-batch gradient descent \cite{watt2020machine}, the process begins by dividing the dataset into temporal segments, each consisting of four sequential time steps. This segmentation is performed for every individual time step within the dataset. Following this, the collection of temporal segments is thoroughly shuffled to ensure randomness. Subsequently, at each training iteration, a subset of these shuffled segments is randomly chosen to form the batch denoted by $S$ in \eqref{equ:rec} and \eqref{equ:jerk}, which is then utilized to evaluate the loss \eqref{equ:stageIloss}.

Once the autoencoder is trained, we feed all snapshots across all trajectories through the encoder, yielding a dataset of solution trajectories in latent space that serves as the foundational input for the subsequent training stage.

\subsubsection{Stage II: Neural ODE Training}
\label{method:stage2}

Following the preparation of the latent dataset, the second stage shifts the focus towards the neural ODE, aiming to learn the continuous latent dynamics. The latent trajectories obtained from the previous stage, each corresponding to a series of snapshots from \(t=0\) to \(t=T\), are utilized to train the neural ODE.

In this stage, the objective is to optimize the neural ODE to accurately model the evolution of latent states over time. The loss function for this stage, \(L_{ODE}\), is primarily designed to minimize the residual between the predicted and actual latent states across all snapshots in the trajectory. Mathematically, it can be expressed as
\begin{equation}
    L_{ODE} = \sum_{t=0}^{T} \| \hat{\boldsymbol{z}}(t) - \boldsymbol{z}(t) \|_2^2,
\end{equation}
where \(\hat{\boldsymbol{z}}(t)\) and \(\boldsymbol{z}(t)\) are the predicted and actual latent vectors at time \(t\), respectively.

In this stage, the mini-batch optimization process is initiated by randomly shuffling the entire latent dataset at the trajectory level to ensure variability and prevent order bias. For each mini-batch, a random collection of trajectories is selected for training purposes. This selection is crucial because the training of neural Ordinary Differential Equations (ODEs) requires the inclusion of complete trajectories to accurately model the dynamical systems.

\subsection{Inference Scheme: Continuous Interpolation in Space and Time}
\label{method:inference}

One of the most distinctive features of our proposed model is its inherent ability to facilitate continuous interpolation across both spatial and temporal dimensions during the inference/testing stage. This capability stems from the integration of two components: the INR-based decoder and the neural ODE framework.

Following the SIREN architechture \cite{sitzmann2020implicit}, the INR-based decoder employs sinewave-activated neural layers for global approximation, inherently offering spatial interpolation capabilities. With a given latent representation \(\boldsymbol{z}(t)\) and spatial coordinates \(\boldsymbol{x}\), the decoder adeptly reconstructs the continuous state \(u(\boldsymbol{x}, t)\). This capability allows it to generate solutions for any spatial point within \(\Omega\), irrespective of its presence in the sampling grid $\{\boldsymbol{x}_i\}_{i=0}^N$. As a result, the model can accurately predict solutions at non-grid spatial locations while retaining essential spatial patterns. This adaptability ensures that the model, even when trained on low-resolution data, can produce high-resolution outputs, offering scalable super-resolution abilities without any architectural alterations.

Temporally, the continuous nature of the neural ODE facilitates interpolation at any time $t$ within the interval $[0, T]$. Discrete-time models such as autoregressive methods in DeepONet and FNO necessitate fixed-time step sizes, thereby limiting their interpolation capabilities. However, owing to the continuous modeling of latent dynamics, the proposed method supports inference at any time \(t\), producing the corresponding latent representation \(\boldsymbol{z}(t)\) and continuous state \(u(\boldsymbol{x}, t)\). This ensures that the model is not merely bound to predicting discrete snapshots in time but can adaptively provide solutions at any time granularity, offering a profound advantage for dynamical systems with multiscale temporal behavior.

In essence, by unifying the power of the INR-based decoder with the flexibility of neural ODEs, our model effortlessly supports spatially and temporally continuous predictions, allowing for a high-fidelity representation of spatiotemporal systems at any desired granularity in both space and time.

\section{Discussion}
The introduced jerk regularization serves as a fundamental physical constraint to enforce smoothness within latent dynamics, thereby enhancing the efficacy of data-driven temporal modeling. This regularization technique, incorporated as a soft constraint during the autoencoder training process, effectively augments the smoothness and sparsity of solution trajectories in the latent space. Beyond facilitating faster convergence rates, it also assists in identifying the intrinsic latent coordinates governing complex systems. While our demonstrations primarily showcase its effectiveness in the context of autoencoders, it is important to emphasize the wide applicability of jerk regularization in the field of data-driven ROMs with compressed latent spaces. Notably, the notion of enforcing smoothness can be seamlessly extended to other architectures, such as the latent space of U-Net and the hidden states of Recurrent Neural Networks (RNNs) or Long Short-Term Memory (LSTM) networks.

It is important to acknowledge a limitation within the current framework, namely, its restriction to fixed input resolutions for the encoder. This constraint can be addressed by adopting an autodecoder approach \cite{wan2023evolve, pan2023neural}, wherein the encoder is removed, and the latent vector can be learned with a limited number of optimization steps using only the decoder. This adaptation not only mitigates the input resolution constraint but also allows for the incorporation of partial observations as inputs. Instead of relying solely on complete input observations, the model can be effectively trained with partial input information, further expanding its utility and versatility. In future work, more challenging examples and more architectures will be included to demonstrate the universality of jerk regularization. 







\bibliographystyle{unsrt}  
\bibliography{references}

\begin{thebibliography}{10}

\bibitem{bi2023accurate}
Kaifeng Bi, Lingxi Xie, Hengheng Zhang, Xin Chen, Xiaotao Gu, and Qi~Tian.
\newblock Accurate medium-range global weather forecasting with 3d neural
  networks.
\newblock {\em Nature}, 619(7970):533--538, 2023.

\bibitem{brunton2020machine}
Steven~L Brunton, Bernd~R Noack, and Petros Koumoutsakos.
\newblock Machine learning for fluid mechanics.
\newblock {\em Annual review of fluid mechanics}, 52:477--508, 2020.

\bibitem{alber2019integrating}
Mark Alber, Adrian Buganza~Tepole, William~R Cannon, Suvranu De, Salvador
  Dura-Bernal, Krishna Garikipati, George Karniadakis, William~W Lytton, Paris
  Perdikaris, Linda Petzold, et~al.
\newblock Integrating machine learning and multiscale modeling—perspectives,
  challenges, and opportunities in the biological, biomedical, and behavioral
  sciences.
\newblock {\em NPJ digital medicine}, 2(1):115, 2019.

\bibitem{mozaffar2022mechanistic}
Mojtaba Mozaffar, Shuheng Liao, Xiaoyu Xie, Sourav Saha, Chanwook Park, Jian
  Cao, Wing~Kam Liu, and Zhengtao Gan.
\newblock Mechanistic artificial intelligence (mechanistic-ai) for modeling,
  design, and control of advanced manufacturing processes: Current state and
  perspectives.
\newblock {\em Journal of Materials Processing Technology}, 302:117485, 2022.

\bibitem{brunton2016discovering}
Steven~L Brunton, Joshua~L Proctor, and J~Nathan Kutz.
\newblock Discovering governing equations from data by sparse identification of
  nonlinear dynamical systems.
\newblock {\em Proceedings of the national academy of sciences},
  113(15):3932--3937, 2016.

\bibitem{xie2022data}
Xiaoyu Xie, Arash Samaei, Jiachen Guo, Wing~Kam Liu, and Zhengtao Gan.
\newblock Data-driven discovery of dimensionless numbers and governing laws
  from scarce measurements.
\newblock {\em Nature communications}, 13(1):7562, 2022.

\bibitem{ren2022phycrnet}
Pu~Ren, Chengping Rao, Yang Liu, Jian-Xun Wang, and Hao Sun.
\newblock Phycrnet: Physics-informed convolutional-recurrent network for
  solving spatiotemporal pdes.
\newblock {\em Computer Methods in Applied Mechanics and Engineering},
  389:114399, 2022.

\bibitem{rao2023encoding}
Chengping Rao, Pu~Ren, Qi~Wang, Oral Buyukozturk, Hao Sun, and Yang Liu.
\newblock Encoding physics to learn reaction-diffusion processes.
\newblock {\em Nature Machine Intelligence}, pages 1--15, 2023.

\bibitem{pfaff2020learning}
Tobias Pfaff, Meire Fortunato, Alvaro Sanchez-Gonzalez, and Peter~W Battaglia.
\newblock Learning mesh-based simulation with graph networks.
\newblock {\em arXiv preprint arXiv:2010.03409}, 2020.

\bibitem{li2020fourier}
Zongyi Li, Nikola Kovachki, Kamyar Azizzadenesheli, Burigede Liu, Kaushik
  Bhattacharya, Andrew Stuart, and Anima Anandkumar.
\newblock Fourier neural operator for parametric partial differential
  equations.
\newblock {\em arXiv preprint arXiv:2010.08895}, 2020.

\bibitem{lu2021learning}
Lu~Lu, Pengzhan Jin, Guofei Pang, Zhongqiang Zhang, and George~Em Karniadakis.
\newblock Learning nonlinear operators via deeponet based on the universal
  approximation theorem of operators.
\newblock {\em Nature machine intelligence}, 3(3):218--229, 2021.

\bibitem{pathak2022fourcastnet}
Jaideep Pathak, Shashank Subramanian, Peter Harrington, Sanjeev Raja, Ashesh
  Chattopadhyay, Morteza Mardani, Thorsten Kurth, David Hall, Zongyi Li, Kamyar
  Azizzadenesheli, et~al.
\newblock Fourcastnet: A global data-driven high-resolution weather model using
  adaptive fourier neural operators.
\newblock {\em arXiv preprint arXiv:2202.11214}, 2022.

\bibitem{seidman2022nomad}
Jacob Seidman, Georgios Kissas, Paris Perdikaris, and George~J Pappas.
\newblock Nomad: Nonlinear manifold decoders for operator learning.
\newblock {\em Advances in Neural Information Processing Systems},
  35:5601--5613, 2022.

\bibitem{tripura2023wavelet}
Tapas Tripura and Souvik Chakraborty.
\newblock Wavelet neural operator for solving parametric partial differential
  equations in computational mechanics problems.
\newblock {\em Computer Methods in Applied Mechanics and Engineering},
  404:115783, 2023.

\bibitem{kissas2022learning}
Georgios Kissas, Jacob~H Seidman, Leonardo~Ferreira Guilhoto, Victor~M
  Preciado, George~J Pappas, and Paris Perdikaris.
\newblock Learning operators with coupled attention.
\newblock {\em The Journal of Machine Learning Research}, 23(1):9636--9698,
  2022.

\bibitem{brandstetter2022message}
Johannes Brandstetter, Daniel Worrall, and Max Welling.
\newblock Message passing neural pde solvers.
\newblock {\em arXiv preprint arXiv:2202.03376}, 2022.

\bibitem{cohen2015approximation}
Albert Cohen and Ronald DeVore.
\newblock Approximation of high-dimensional parametric pdes.
\newblock {\em Acta Numerica}, 24:1--159, 2015.

\bibitem{benner2015survey}
Peter Benner, Serkan Gugercin, and Karen Willcox.
\newblock A survey of projection-based model reduction methods for parametric
  dynamical systems.
\newblock {\em SIAM review}, 57(4):483--531, 2015.

\bibitem{rowley2017model}
Clarence~W Rowley and Scott~TM Dawson.
\newblock Model reduction for flow analysis and control.
\newblock {\em Annual Review of Fluid Mechanics}, 49:387--417, 2017.

\bibitem{lee2020model}
Kookjin Lee and Kevin~T Carlberg.
\newblock Model reduction of dynamical systems on nonlinear manifolds using
  deep convolutional autoencoders.
\newblock {\em Journal of Computational Physics}, 404:108973, 2020.

\bibitem{chen2022crom}
Peter~Yichen Chen, Jinxu Xiang, Dong~Heon Cho, Yue Chang, GA~Pershing,
  Henrique~Teles Maia, Maurizio~M Chiaramonte, Kevin~Thomas Carlberg, and Eitan
  Grinspun.
\newblock Crom: Continuous reduced-order modeling of pdes using implicit neural
  representations.
\newblock In {\em The Eleventh International Conference on Learning
  Representations}, 2022.

\bibitem{yin2023continuous}
Yuan Yin, Matthieu Kirchmeyer, Jean-Yves Franceschi, Alain Rakotomamonjy, and
  Patrick Gallinari.
\newblock Continuous pde dynamics forecasting with implicit neural
  representations.
\newblock In {\em The Eleventh International Conference on Learning
  Representations}, 2023.

\bibitem{sholokhov2023physics}
Aleksei Sholokhov, Yuying Liu, Hassan Mansour, and Saleh Nabi.
\newblock Physics-informed neural ode (pinode): embedding physics into models
  using collocation points.
\newblock {\em Scientific Reports}, 13(1):10166, 2023.

\bibitem{he2023glasdi}
Xiaolong He, Youngsoo Choi, William~D Fries, Jonathan~L Belof, and Jiun-Shyan
  Chen.
\newblock Glasdi: Parametric physics-informed greedy latent space dynamics
  identification.
\newblock {\em Journal of Computational Physics}, page 112267, 2023.

\bibitem{wan2023evolve}
Zhong~Yi Wan, Leonardo Zepeda-N{\'u}{\~n}ez, Anudhyan Boral, and Fei Sha.
\newblock Evolve smoothly, fit consistently: Learning smooth latent dynamics
  for advection-dominated systems.
\newblock {\em arXiv preprint arXiv:2301.10391}, 2023.

\bibitem{serrano2023operator}
Louis Serrano, Lise~Le Boudec, Armand~Kassa{\"\i} Koupa{\"\i}, Thomas~X Wang,
  Yuan Yin, Jean-No{\"e}l Vittaut, and Patrick Gallinari.
\newblock Operator learning with neural fields: Tackling pdes on general
  geometries.
\newblock {\em arXiv preprint arXiv:2306.07266}, 2023.

\bibitem{raut2023arousal}
Ryan~V Raut, Zachary~P Rosenthal, Xiaodan Wang, Hanyang Miao, Zhanqi Zhang,
  Jin-Moo Lee, Marcus~E Raichle, Adam~Q Bauer, Steven~L Brunton, Bingni~W
  Brunton, et~al.
\newblock Arousal as a universal embedding for spatiotemporal brain dynamics.
\newblock {\em bioRxiv}, pages 2023--11, 2023.

\bibitem{otto2023learning}
Samuel~E Otto, Gregory~R Macchio, and Clarence~W Rowley.
\newblock Learning nonlinear projections for reduced-order modeling of
  dynamical systems using constrained autoencoders.
\newblock {\em Chaos: An Interdisciplinary Journal of Nonlinear Science},
  33(11), 2023.

\bibitem{chen2024nonlinear}
Peiyi Chen, Tianchen Hu, and Johann Guilleminot.
\newblock A nonlinear-manifold reduced-order model and operator learning for
  partial differential equations with sharp solution gradients.
\newblock {\em Computer Methods in Applied Mechanics and Engineering},
  419:116684, 2024.

\bibitem{moore1981principal}
Bruce Moore.
\newblock Principal component analysis in linear systems: Controllability,
  observability, and model reduction.
\newblock {\em IEEE transactions on automatic control}, 26(1):17--32, 1981.

\bibitem{ravindran2000reduced}
Sivaguru~S Ravindran.
\newblock A reduced-order approach for optimal control of fluids using proper
  orthogonal decomposition.
\newblock {\em International journal for numerical methods in fluids},
  34(5):425--448, 2000.

\bibitem{holmes2012turbulence}
Philip Holmes, John Lumley, Gahl Berkooz, and Clarence Rowley.
\newblock {\em Turbulence, coherent structures, dynamical systems and
  symmetry}.
\newblock Cambridge university press, 2012.

\bibitem{willcox2002balanced}
Karen Willcox and Jaime Peraire.
\newblock Balanced model reduction via the proper orthogonal decomposition.
\newblock {\em AIAA journal}, 40(11):2323--2330, 2002.

\bibitem{rowley2005model}
Clarence~W Rowley.
\newblock Model reduction for fluids, using balanced proper orthogonal
  decomposition.
\newblock {\em International Journal of Bifurcation and Chaos},
  15(03):997--1013, 2005.

\bibitem{chinesta2011overview}
Francisco Chinesta, Amine Ammar, Adrien Leygue, and Roland Keunings.
\newblock An overview of the proper generalized decomposition with applications
  in computational rheology.
\newblock {\em Journal of Non-Newtonian Fluid Mechanics}, 166(11):578--592,
  2011.

\bibitem{pinkus2012n}
Allan Pinkus.
\newblock {\em N-widths in Approximation Theory}, volume~7.
\newblock Springer Science \& Business Media, 2012.

\bibitem{peherstorfer2022breaking}
Benjamin Peherstorfer.
\newblock Breaking the kolmogorov barrier with nonlinear model reduction.
\newblock {\em Notices of the American Mathematical Society}, 69(5):725--733,
  2022.

\bibitem{dissanayake1994neural}
MWMG Dissanayake and Nhan Phan-Thien.
\newblock Neural-network-based approximations for solving partial differential
  equations.
\newblock {\em communications in Numerical Methods in Engineering},
  10(3):195--201, 1994.

\bibitem{lagaris1998artificial}
Isaac~E Lagaris, Aristidis Likas, and Dimitrios~I Fotiadis.
\newblock Artificial neural networks for solving ordinary and partial
  differential equations.
\newblock {\em IEEE transactions on neural networks}, 9(5):987--1000, 1998.

\bibitem{raissi2019physics}
Maziar Raissi, Paris Perdikaris, and George~E Karniadakis.
\newblock Physics-informed neural networks: A deep learning framework for
  solving forward and inverse problems involving nonlinear partial differential
  equations.
\newblock {\em Journal of Computational physics}, 378:686--707, 2019.

\bibitem{cai2021physics}
Shengze Cai, Zhiping Mao, Zhicheng Wang, Minglang Yin, and George~Em
  Karniadakis.
\newblock Physics-informed neural networks (pinns) for fluid mechanics: A
  review.
\newblock {\em Acta Mechanica Sinica}, 37(12):1727--1738, 2021.

\bibitem{karniadakis2021physics}
George~Em Karniadakis, Ioannis~G Kevrekidis, Lu~Lu, Paris Perdikaris, Sifan
  Wang, and Liu Yang.
\newblock Physics-informed machine learning.
\newblock {\em Nature Reviews Physics}, 3(6):422--440, 2021.

\bibitem{mowlavi2023optimal}
Saviz Mowlavi and Saleh Nabi.
\newblock Optimal control of pdes using physics-informed neural networks.
\newblock {\em Journal of Computational Physics}, 473:111731, 2023.

\bibitem{zhao2023physics}
Xiaoyu Zhao, Zhiqiang Gong, Yunyang Zhang, Wen Yao, and Xiaoqian Chen.
\newblock Physics-informed convolutional neural networks for temperature field
  prediction of heat source layout without labeled data.
\newblock {\em Engineering Applications of Artificial Intelligence},
  117:105516, 2023.

\bibitem{li2021physics}
Zongyi Li, Hongkai Zheng, Nikola Kovachki, David Jin, Haoxuan Chen, Burigede
  Liu, Kamyar Azizzadenesheli, and Anima Anandkumar.
\newblock Physics-informed neural operator for learning partial differential
  equations.
\newblock {\em arXiv preprint arXiv:2111.03794}, 2021.

\bibitem{wang2021learning}
Sifan Wang, Hanwen Wang, and Paris Perdikaris.
\newblock Learning the solution operator of parametric partial differential
  equations with physics-informed deeponets.
\newblock {\em Science advances}, 7(40):eabi8605, 2021.

\bibitem{goswami2023physics}
Somdatta Goswami, Aniruddha Bora, Yue Yu, and George~Em Karniadakis.
\newblock Physics-informed deep neural operator networks.
\newblock In {\em Machine Learning in Modeling and Simulation: Methods and
  Applications}, pages 219--254. Springer, 2023.

\bibitem{navaneeth2024physics}
N~Navaneeth, Tapas Tripura, and Souvik Chakraborty.
\newblock Physics informed wno.
\newblock {\em Computer Methods in Applied Mechanics and Engineering},
  418:116546, 2024.

\bibitem{ling2016machine}
Julia Ling, Reese Jones, and Jeremy Templeton.
\newblock Machine learning strategies for systems with invariance properties.
\newblock {\em Journal of Computational Physics}, 318:22--35, 2016.

\bibitem{ling2016reynolds}
Julia Ling, Andrew Kurzawski, and Jeremy Templeton.
\newblock Reynolds averaged turbulence modelling using deep neural networks
  with embedded invariance.
\newblock {\em Journal of Fluid Mechanics}, 807:155--166, 2016.

\bibitem{villar2021scalars}
Soledad Villar, David~W Hogg, Kate Storey-Fisher, Weichi Yao, and Ben
  Blum-Smith.
\newblock Scalars are universal: Equivariant machine learning, structured like
  classical physics.
\newblock {\em Advances in Neural Information Processing Systems},
  34:28848--28863, 2021.

\bibitem{gabel2023learning}
Alex Gabel, Victoria Klein, Riccardo Valperga, Jeroen~SW Lamb, Kevin Webster,
  Rick Quax, and Efstratios Gavves.
\newblock Learning lie group symmetry transformations with neural networks.
\newblock In {\em Topological, Algebraic and Geometric Learning Workshops
  2023}, pages 50--59. PMLR, 2023.

\bibitem{choi2023symmetry}
Hee-Sun Choi, Yonggyun Yu, and Hogeon Seo.
\newblock Symmetry-informed surrogates with data-free constraint for real-time
  acoustic wave propagation.
\newblock {\em Applied Acoustics}, 214:109686, 2023.

\bibitem{hansen2023learning}
Derek Hansen, Danielle~C Maddix, Shima Alizadeh, Gaurav Gupta, and Michael~W
  Mahoney.
\newblock Learning physical models that can respect conservation laws.
\newblock {\em arXiv preprint arXiv:2302.11002}, 2023.

\bibitem{desai2024trade}
Saaketh Desai, Ankit Shrivastava, Marta D’Elia, Habib~N Najm, and R{\'e}mi
  Dingreville.
\newblock Trade-offs in the latent representation of microstructure evolution.
\newblock {\em Acta Materialia}, 263:119514, 2024.

\bibitem{pan2023neural}
Shaowu Pan, Steven~L Brunton, and J~Nathan Kutz.
\newblock Neural implicit flow: a mesh-agnostic dimensionality reduction
  paradigm of spatio-temporal data.
\newblock {\em Journal of Machine Learning Research}, 24(41):1--60, 2023.

\bibitem{gupta2022towards}
Jayesh~K Gupta and Johannes Brandstetter.
\newblock Towards multi-spatiotemporal-scale generalized pde modeling.
\newblock {\em arXiv preprint arXiv:2209.15616}, 2022.

\bibitem{mialon2023self}
Gr{\'e}goire Mialon, Quentin Garrido, Hannah Lawrence, Danyal Rehman, Yann
  LeCun, and Bobak~T Kiani.
\newblock Self-supervised learning with lie symmetries for partial differential
  equations.
\newblock {\em arXiv preprint arXiv:2307.05432}, 2023.

\bibitem{mavi2023unsupervised}
Arda Mavi, Ali~Can Bekar, Ehsan Haghighat, and Erdogan Madenci.
\newblock An unsupervised latent/output physics-informed convolutional-lstm
  network for solving partial differential equations using peridynamic
  differential operator.
\newblock {\em Computer Methods in Applied Mechanics and Engineering},
  407:115944, 2023.

\bibitem{champion2019data}
Kathleen Champion, Bethany Lusch, J~Nathan Kutz, and Steven~L Brunton.
\newblock Data-driven discovery of coordinates and governing equations.
\newblock {\em Proceedings of the National Academy of Sciences},
  116(45):22445--22451, 2019.

\bibitem{he2016deep}
Kaiming He, Xiangyu Zhang, Shaoqing Ren, and Jian Sun.
\newblock Deep residual learning for image recognition.
\newblock In {\em Proceedings of the IEEE conference on computer vision and
  pattern recognition}, pages 770--778, 2016.

\bibitem{he2015delving}
Kaiming He, Xiangyu Zhang, Shaoqing Ren, and Jian Sun.
\newblock Delving deep into rectifiers: Surpassing human-level performance on
  imagenet classification.
\newblock In {\em Proceedings of the IEEE international conference on computer
  vision}, pages 1026--1034, 2015.

\bibitem{chen2018neural}
Ricky~TQ Chen, Yulia Rubanova, Jesse Bettencourt, and David~K Duvenaud.
\newblock Neural ordinary differential equations.
\newblock {\em Advances in neural information processing systems}, 31, 2018.

\bibitem{hornik1989multilayer}
Kurt Hornik, Maxwell Stinchcombe, and Halbert White.
\newblock Multilayer feedforward networks are universal approximators.
\newblock {\em Neural networks}, 2(5):359--366, 1989.

\bibitem{elfwing2018sigmoid}
Stefan Elfwing, Eiji Uchibe, and Kenji Doya.
\newblock Sigmoid-weighted linear units for neural network function
  approximation in reinforcement learning.
\newblock {\em Neural networks}, 107:3--11, 2018.

\bibitem{park2019deepsdf}
Jeong~Joon Park, Peter Florence, Julian Straub, Richard Newcombe, and Steven
  Lovegrove.
\newblock Deepsdf: Learning continuous signed distance functions for shape
  representation.
\newblock In {\em Proceedings of the IEEE/CVF conference on computer vision and
  pattern recognition}, pages 165--174, 2019.

\bibitem{sitzmann2020implicit}
Vincent Sitzmann, Julien Martel, Alexander Bergman, David Lindell, and Gordon
  Wetzstein.
\newblock Implicit neural representations with periodic activation functions.
\newblock {\em Advances in neural information processing systems},
  33:7462--7473, 2020.

\bibitem{watt2020machine}
Jeremy Watt, Reza Borhani, and Aggelos~K Katsaggelos.
\newblock {\em Machine learning refined: Foundations, algorithms, and
  applications}.
\newblock Cambridge University Press, 2020.

\end{thebibliography}

\end{document}